\pdfoutput=1
\pdfoutput=1
\pdfoutput=1
\pdfoutput=1
\pdfoutput=1

\documentclass[runningheads]{llncs}
\usepackage{graphicx}
\usepackage{mathrsfs}
\usepackage{amsmath,amssymb,amsfonts}
\usepackage{algorithm} 
\usepackage{algorithmic}
\usepackage{caption}
\usepackage{multirow}
\usepackage{xcolor}

\begin{document}
\title{Improving Many-Objective Evolutionary Algorithms by Means of Edge-Rotated Cones\thanks{This work is part of the research programme Smart Industry SI2016 with project name CIMPLO and project number 15465, which is (partly) financed by the Netherlands Organisation for Scientific Research (NWO).}}

%
\titlerunning{Improving MOEAs by Means of Edge-Rotated Cones}

%

\author{Yali Wang \and
Andr{\'e} Deutz \and 
Thomas B{\"a}ck \and
Michael Emmerich
}
\authorrunning{Yali Wang et al.}
%
\institute{Leiden Institute of Advanced Computer Science, Leiden University, Niels Bohrweg 1, 2333CA Leiden, The Netherlands\\
\email{y.wang@liacs.leidenuniv.nl}\\
}

\maketitle              
\begin{abstract}
Given a point in $m$-dimensional objective space, any $\varepsilon$-ball of a point can be partitioned into the incomparable, the dominated and dominating region. The ratio between the size of the incomparable region, and the dominated (and dominating) region decreases proportionally to $1/2^{m-1}$, i.e., the volume of the Pareto dominating orthant as compared to all other volumes. Due to this reason, it gets increasingly unlikely that dominating points can be found by random, isotropic mutations. As a remedy to stagnation of search in many objective optimization, in this paper, we suggest to enhance the Pareto dominance order by involving an obtuse convex dominance cone in the convergence phase of an evolutionary optimization algorithm. We propose edge-rotated cones as generalizations of Pareto dominance cones for which the opening angle can be controlled by a single parameter only. The approach is integrated in several state-of-the-art multi-objective evolutionary algorithms (MOEAs) and tested on benchmark problems with four, five, six and eight objectives. Computational experiments demonstrate the ability of these edge-rotated cones to improve the performance of MOEAs on many-objective optimization problems.

\end{abstract}
\section{Introduction}
Multi-objective evolutionary algorithms (MOEAs) have been successfully used in the application area of multi-objective optimization due to their ability to approximate the entire Pareto front in a single run. The Pareto dominance relation, as the most commonly adopted ranking method, plays an essential role in many MOEAs because Pareto dominance is used to compare solutions even when different selection mechanisms are employed in different categories of MOEAs. The well-known NSGA-II \cite{deb2002fast} is a Pareto dominance-based MOEA, using Pareto non-dominated sorting as the first ranking criterion and crowding distance to promote diversity in the population. DI-MOEA \cite{dimoea} is an indicator-based MOEA, it employs the non-dominated sorting as the first ranking criterion and a diversity indicator as the second criterion, which is the Euclidean distance based geometric mean gap indicator. It has been shown to be invariant to the shape of the Pareto front and can achieve evenly spread Pareto front approximations. The NSGA-III \cite{deb2013evolutionary} is an extension of NSGA-II and it is a decomposition-based MOEA. It employs the Pareto non-dominated sorting to partition the population into a number of fronts, but replaces the crowding distance operator with a clustering operator based on a set of reference points. 

Although the Pareto dominance relation usually works well on multi-objective problems with two or three objectives, its ability is often severely degraded when handling many-objective problems (MaOPs) where more than three objectives need to be optimized simultaneously. 
One major reason of its performance deterioration in many-objective optimization is that individuals are not likely to be dominated by others. Given a point in $m$-dimensional objective space, any $\varepsilon$-ball of a point can be partitioned into the incomparable, the dominated and dominating region. The ratio between the size of the incomparable region, and the dominated (and dominating) region decreases proportionally to $1/2^{m-1}$, i.e., the volume of the Pareto dominating orthant as compared to all other volumes. Due to this reason, it gets increasingly unlikely that dominating points can be found by random, isotropic mutations and classical algorithms do not converge to the Pareto front.
The straightforward attempt to overcome the weakness is to use a large population. However, the use of a large population causes other issues. Firstly, the computing time of MOEAs drastically increases because of the increase of the population size. Secondly, the use of a large population size severely degrades the search ability of some MOEAs, (e.g., NSGA-II) \cite{ishibuchi2009evolutionary}. 
Instead, we propose to extend the Pareto dominance order during the convergence phase by involving the cone order from a convex obtuse dominance cone. The new cone is implemented by rotating the edges of the standard Pareto cone by means of a single parameter. In this way, an individual can dominate larger space, thus, a gradient towards dominating solutions can be followed using relatively small population sizes. 

The structure of this paper is as follows. After discussing related work (Sect.~\ref{sec:relatedwork}), Sect.~\ref{sec:algorithm} describes the edge-rotated cone dominance approach. Sect.~\ref{sec:experiment} presents a comparative analysis. Finally, Sect.~\ref{sec:conclusion} concludes the paper.

\vspace{-0.3cm}
\section{Related Work}
\label{sec:relatedwork}

The Pareto dominance relationship is the most commonly adopted ranking method in multi-objective optimization. However, with the increase of the number of objectives, the convergence ability of MOEAs based on Pareto dominance degrades significantly \cite{khare2003performance}. Recently, some researchers have proposed the use of relaxed forms of Pareto dominance as a way of regulating convergence of MOEAs. Under these relaxed definitions, a solution has a higher chance to be dominated by other solutions and the selection pressure toward the Pareto front is increased.

\begin{definition}
(Pareto dominance) An objective vector $y^{(1)}\in \mathbb{R}^m$ is said to dominate another objective vector $y^{(2)} \in \mathbb{R}^m$  (denoted by $y^{(1)} \prec_{pareto} y^{(2)}$) if and only if: $y_i^{(1)} \leq y_i^{(2)} ~\forall i=1,\dots,m$ and $\exists i \in \{1,\dots,m\}: y_i^{(1)} < y_i^{(2)}$.
\end{definition}

Ikeda et al. proposed $\alpha$-dominance \cite{ikeda2001failure} to deal with dominance resistant solutions (DRSs), which are solutions far from the Pareto front but are hardly dominated. In $\alpha$-dominance, the upper and lower bounds of trade-off rates between two objectives $f_i$ and $f_j$, i.e., $\alpha_{ij}$ and $\alpha_{ji}$, are pre-defined. Before judging the dominance relations between two individuals $y$ and $y'$ in the population, the following definition is considered:
$g_i(y, y') := f_i(y)-f_i(y')+\sum_{j\neq i}^{M}\alpha_{ij}(f_j(y)-f_{j}(y'))$. Solution $y$ dominates solution $y'$ if and only if $\forall i\in\{1,...,m\}: g_i(y,y')\leq 0$ and $\exists i \in\{1,...,m\}: g_i(y,y')< 0$. Using $\alpha$-dominance allows a solution to dominate another if it is slightly inferior to the other in one objective, but largely superior in other objectives by setting lower and upper bounds of trade-off rates between objectives.

Laumanns et al. proposed the concept of $\epsilon$-dominance \cite{laumanns2002combining}. Given two solutions $y$, $y'\in \mathbb{R}^m$, and $\epsilon > 0$, $y$ is said to $\epsilon$-dominant $y'$ if and only if $\forall i \in \{1,...,m\}$: $y_i-\epsilon\leq y_i^{'}$.
Cone $\epsilon$-dominance \cite{batista2011pareto} has been proposed by Batista et al. to improve $\epsilon$-dominance which may eliminate viable solutions. It introduces a parameter $k$ ($k\in [0,1)$) to control the shape of the dominance area of a solution using cones.
Cone-dominance is also prominently used in multi-criteria decision making (MCDM), in order to formulate user preferences \cite{wiecek2007}.

Sato et al. proposed an approach to control the dominance area of solutions (CDAS) \cite{sato2007controlling}. In CDAS, the objective values are modified and the $i$-th objective value of $x$ after modification is defined as:
$\hat{f}_i(x)=\frac{r\cdot\sin{(w_i+S_i\cdot \pi)}}{\sin{(S_i\cdot \pi)}}$,
where $r$ is the norm of $f(x)$, $w_i$ is the declination angle between $f(x)$ and the coordinate axis. The degree of expansion or contraction can be controlled by the parameter $S_i \in[0.25,0.25]$. CDAS controls the aperture of the cone of dominance so that the influence of each point could be increased.

Yang et al. proposed a grid dominance relation \cite{yang2013grid} in the grid-based evolutionary algorithm (GrEA). The grid dominance adds the selection pressure by adopting an adaptive grid construction. It uses grid-based convergence and diversity measurements to compare non-dominated solutions.

Recently, an angle dominance criterion was proposed in \cite{liu2020angle}. It designs a parameter $k$ which works together with the worst point of the current population to control the dominance area of a solution. The angle of a solution (e.g., solution $y$) on one objective (e.g., the $i$th objective), $\alpha_i^y$, is determined by two lines: the $i$th axis; and the line connecting the solution and the farthest point on the $i$th axis in the dominance area. Solution $y$ angle dominates solution $y'$ if and only if $\forall i\in\{1,...,m\}: \alpha_i^y\leq \alpha_i^{y'}$ and $\exists i \in\{1,...,m\}: \alpha_i^y< \alpha_i^{y'}$.

Other than these, the ($1-k$)-based criterion \cite{farina2002optimal} has been considered when addressing MaOPs. After comparing a solution to another and counting the number of objectives where it is better than, the same as, or worse than the other, this criterion uses these numbers to distinguish the relations of domination between solutions.
The $k$-optimality \cite{farina2004fuzzy} is a relation based on the number of improved objectives between two solutions.
The $l$-optimality \cite{zou2008new} not only takes into account the number of improved objective values but also considers the values of improved objective functions, if all objectives have the same importance. The concept of volume dominance was proposed by Le and Landa-Silva \cite{le2007obtaining}. This form of dominance is based on the volume of the objective space that a solution dominates.  

In this paper, we propose the approach of using the edge-rotated cone to enhance the traditional Pareto dominance. The edge-rotated cone can lead to the same dominance relation as $\alpha$-dominance. However, it is interpreted in a more intuitive and geometric way and compared to angle-based method does not require the knowledge of the ideal point or the nadir point. 

\vspace{-0.2cm}
\section{Proposed Algorithm}
\label{sec:algorithm}
\vspace{-0.2cm}
\subsection{Proposed dominance relation}
\begin{figure}[htbp]
\centering
\vspace{-0.8cm}
\begin{minipage}[t]{0.6\textwidth}
\hspace{-0.5cm}
\includegraphics[width=8.3cm]{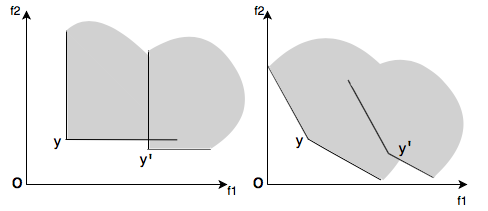}
\caption{Pareto and edge-rotated cone dominance.}
\label{f2}
\end{minipage}
\begin{minipage}[t]{0.38\textwidth}
\hspace{0.8cm}
\includegraphics[width=3.6cm]{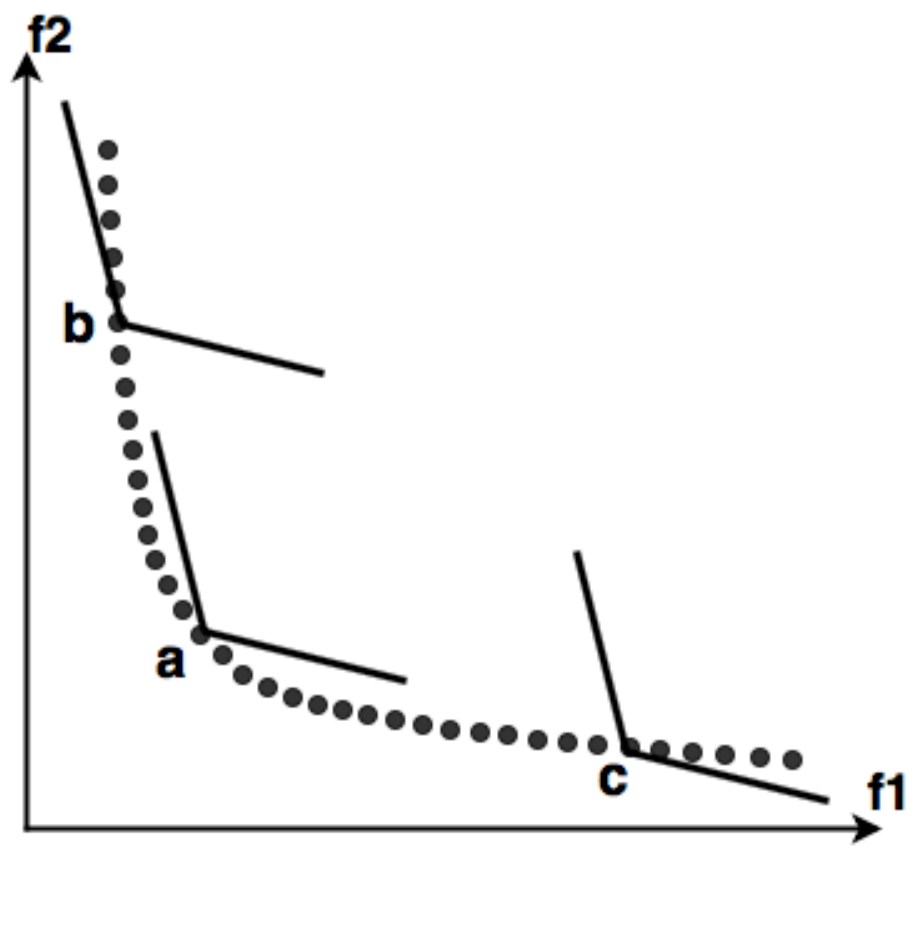}
\caption{Trade-off on PF.}
\label{ff}
\end{minipage}
\end{figure}

The Pareto dominance relation or Pareto order ($\prec_{pareto}$) is a special case of cone orders, which are orders defined on vector spaces. The left image of Figure~\ref{f2} shows an example of applying the Pareto order cone to illustrate the Pareto dominance relation, i.e., $y$ dominates the points in $y \oplus \mathbb{R}^2_{\succ o}$ and $y'$ dominates the points in $y' \oplus \mathbb{R}^2_{\succ o}$. Here, $\mathbb{R}^2_{\succ o}$ is the Pareto order cone and $\oplus$ is the Minkowski sum.

\begin{definition}
(Cone) A set $C$ is a cone if $\lambda w \in C$ for any $w \in C$ and $\forall \lambda > 0$.
\end{definition}

\begin{definition}
(Minkowski Sum) The Minkowski sum (aka
algebraic sum) of two sets $A \in \mathbb{R}^m$ and $B \in \mathbb{R}^m$ is defined as $A\oplus B :=\{a+b~|~a\in A \land b\in B\}$.
\end{definition}


\vspace{-0.5cm}

In an MOEA, if a solution can dominate more area based on the adopted dominance relation, the algorithm is capable of exploring more solutions and hence accelerating convergence. To this end, we widen the angle of the Pareto order cone and generate the cone which can dominate a larger area. Given a linearly independent vector set $\{w_1, w_2, \dots, w_m\}$, a cone can be generated in $m$-dimensional space.

\begin{definition}
\label{def:gecone}
(Generated $m$-dimensional cone) The cone generated by the vectors $w_1, w_2,\dots, w_m$ is the set $C = \{z:z=\lambda_1 w_1 + \lambda_2 w_2 + \dots + \lambda_m w_m, \forall\lambda_1, \lambda_2,\dots,\\ \lambda_m \geq0, \lambda \neq 0\}$; $w_1$, \dots, $w_m$ are linearly independent.
\end{definition}

To be specific, the Pareto order cone is widened by rotating the edges of the standard Pareto order cone around the origin towards the outside. For example, in two-dimensional space, the Pareto order cone is the cone generated by two axes which support an angle of 90$^{\circ}$. By rotating two axes towards the opposite direction around the origin, the two axes can reach into the second and fourth quadrants respectively and an edge-rotated cone with an angle larger than 90$^{\circ}$ is generated. The right image of Figure~\ref{f2} shows how the dominance relation has been changed when the edge-rotated cone order is applied. In the left image of Figure~\ref{f2}, $y$ and $y'$ are mutually non-dominated by each other because neither of them is in the dominating space of the other point. However, when an edge-rotated cone is adopted in the right image, the point $y'$ is dominated by $y$. We can see that the edge-rotated cones provide a stricter order compared to the Pareto order. They can guide the search towards the Pareto front better as they establish an ordering among the incomparable solutions (with respect to the Pareto order) in the sense that better incomparable solutions are preferred.

When using the edge-rotated cone order in MOEAs, since the concave cones do not give rise to a strict partial order and the non-dominated points in the order generated by acute-angle cones can be dominated in the Pareto order, we restrict ourselves to convex obtuse cones obtained by rotating each edge of the standard Pareto cone towards the outside with an angle of less than 45$^{\circ}$.


\begin{definition}
(Convex Cone) A cone $\mathcal{C}$ is convex if and only if $\forall c_1 \in \mathcal{C}, c_2 \in \mathcal{C}, \forall\alpha ~(0\leq\alpha\leq 1): \alpha c_1+(1-\alpha) c_2\in \mathcal{C}$.
\end{definition}

The approach of widening the standard Pareto cone in $m$-dimensional space ($m>2$) is the same. Each edge of the standard Pareto order cone is rotated by an angle less than 45$^{\circ}$ in the opposite direction of the identity line in the positive orthant. The rotation takes place in the plane determined by the edge and the identity line.
In $m$-dimensional space, the identity line in the positive orthant is the line passing through the origin and the point $(1, ..., 1)$. The new cone composed of the rotated edges can give rise to a new dominance relation.

\vspace{-0.2cm}
\subsection{Implementation and Integration in MOEAs}

In a multi-objective optimization algorithm, solutions that are dominating under the Pareto order are also dominating under the edge-rotated cone order. In this way, it is guaranteed that a minimal element of the edge-rotated cone order is also a minimal element of the Pareto order, and thus algorithms that converge to globally efficient points under the edge-rotated cone order will also converge to globally Pareto efficient points. By using the edge-rotated cone, a solution, especially the solution which is not in the knee region, has a higher chance to be dominated by other solutions. The knee region is the region where the maximum trade-off of objective functions takes place. For the Pareto front in Figure~\ref{ff}, the knee region is where the Pareto surface bulges the most, i.e., the region near solution $a$. When comparing the knee point $a$ with another solution $c$, solution $c$ has a better (i.e., lower) $f2$ value as compared to solution $a$. However, this small improvement leads to a large deterioration in the other objective $f1$. Due to the reason that in the absence of explicitly provided preferences, all objectives are considered equally important, solution $a$, thus, is more preferable than solution $c$. It has been argued in the literature that knee points are the most interesting solutions and preferred solutions \cite{das1999characterizing} - \cite{braun2011preference}. Therefore, although not all globally efficient points might be obtained by the edge-rotated cone orders, the edge-rotated cone orders naturally filter out non-preferred solutions. In Figure~\ref{ff}, when applying the edge-rotated cone, solutions in the knee region can survive, while solutions like $b$ and $c$ are on the flat Pareto surface and are more easily to be dominated. 
\vspace{-0.3cm}
\begin{algorithm}[]
 	\caption{Applying a proper cone order in each iteration.}
    \label{algorithm:angle}
 	\begin{algorithmic}[1]
        \STATE $m \leftarrow$ the number of objectives;
        \STATE $Degree[m]$;  // the rotation angle for each edge of the standard Pareto order;
        \STATE $n\_rank \leftarrow$ Pareto rank number of current population;
        \IF{$n\_rank = 0 $}
        \FOR{{\bf each} $i \in \{ 1, \dots, m\}$}
        \STATE $Degree[i] \leftarrow PI/6$; // rotation angle is $30^{\circ}$
        \ENDFOR
        \ELSE
        \FOR{{\bf each} $i \in \{ 1, \dots, m\}$}
        \STATE $Degree[i] \leftarrow 0$;  // standard Pareto cone
        \ENDFOR
        \ENDIF
 	\end{algorithmic}
\end{algorithm}
\vspace{-0.4cm}

The feature of the edge-rotated cone to eliminate solutions can be appreciated as an advantage especially in the realm of many-objective optimization considering the exponential increase in the number of non-dominated solutions necessary for approximating the entire Pareto front. With the edge-rotated cone, part of the solutions, especially non-preferred solutions, can be excluded. However, this could degrade the diversity of the solution set. Therefore, we propose Algorithm 1 to choose a proper cone order in each iteration of MOEAs in order to promote diversity in addition to convergence. When running an MOEA, the current population is ranked based on the current cone order at the beginning of each iteration; the edge-rotated cone will be adopted only under the condition that all solutions in the current population are mutually non-dominated by each other. In the case that the current population consists of multiple layers, the standard Pareto cone is used (i.e., the rotation angle is 0$^{\circ}$). The underlying idea is when all the solutions are non-dominated with each other, the edge-rotated cone is adopted to enhance the selection pressure, otherwise, the Pareto order cone is used to maintain the diversity of the population.

\begin{figure}
\hspace{-1cm}
\includegraphics[width=5in]{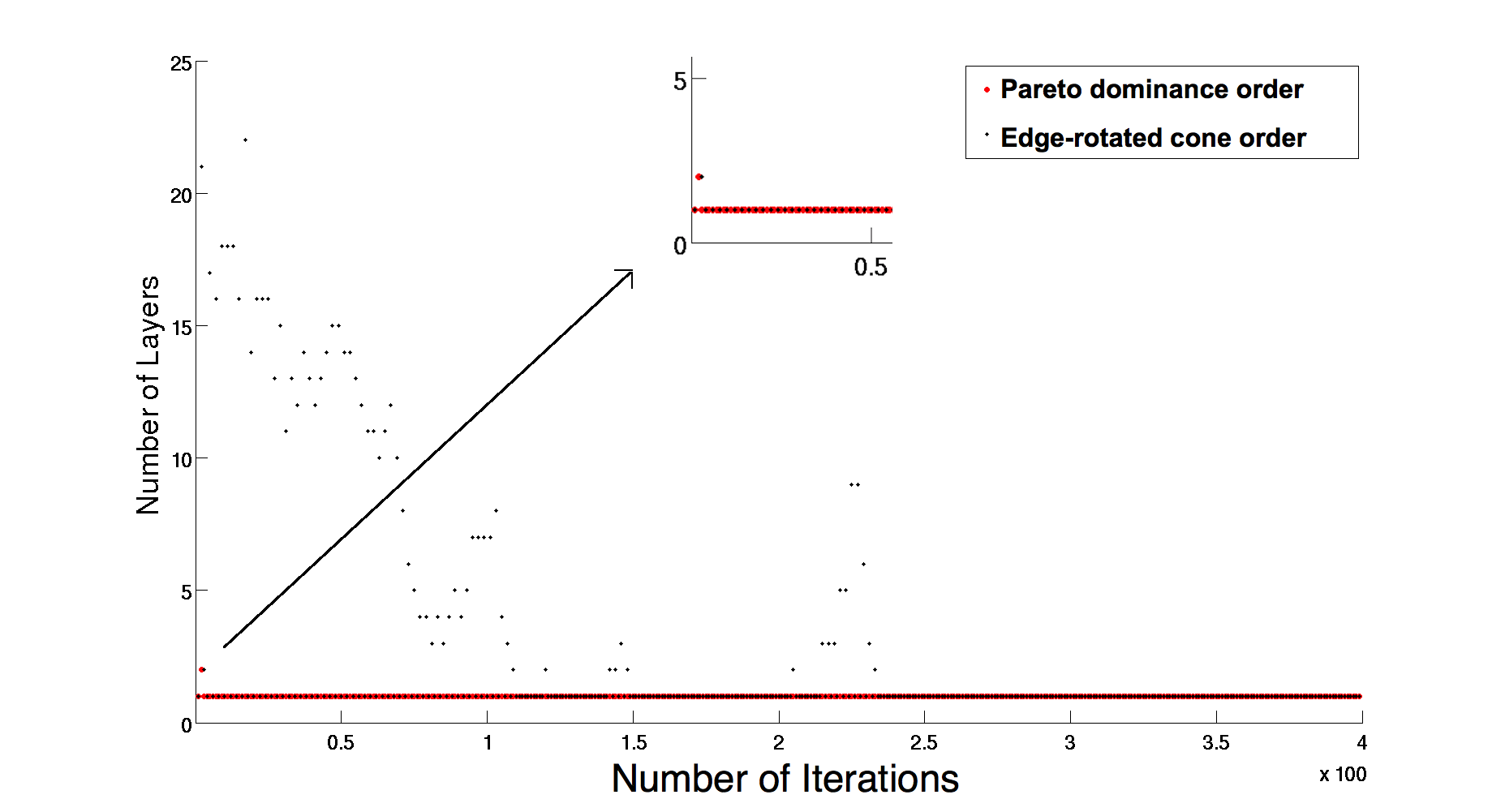}
\caption{The dynamics of the number of layers.} \label{xe}
\vspace{-0.8cm}
\end{figure}

When Algorithm 1 is applied in NSGA-II on the DTLZ1 eight objective problem, Figure~\ref{xe} compares the changes of the number of layers between running NSGA-II using only the Pareto dominance and involving the edge-rotated cone order with a rotation angle of $20^\circ$ within the first $400$ iterations (Population size is 100.). When running the original NSGA-II, except that one point lies at level $2$ (i.e., the number of fronts is two) at the very beginning, the number of layers always remains one, meaning that all solutions in the current population are non-dominated with each other. As a result, the Pareto dominance relation has no effect on parent selection. That is, an individual with a larger crowding distance is always chosen as a parent in the binary tournament selection since all solutions have the same rank. In this manner, the selection pressure toward the Pareto front is severely weakened. However, when the edge-rotated cone is involved, the layering of the population is very noticeable. In this case, an ordering among the incomparable solutions is established and it can guide the search towards the Pareto front better.

Next we derive a criterion by which one can determine whether a point $y' \in \mathbb{R}^2 $ is dominated by a point $y \in \mathbb{R}^2$ with respect to the edge-rotated cone order. 
Let $e_1 := 
\left [
\begin{matrix}
1 \\
0
\end{matrix}
\right]
$
and  $e_2 :=
\left [
\begin{matrix}
0 \\
1
\end{matrix}
\right]
$
be the edges of the two-dimensional standard Pareto cone. Then the edges of the edge-rotated cone by a rotation angle $\alpha$ ( $0 \leq \alpha < \frac{\pi}{4}$) are 
$A e_1$ and $A e_2$, where 
$A =
\left [
\begin{matrix}
\cos (-\alpha) & \frac{\sin(-\alpha)}{\sqrt{2-1}} \\
\frac{\sin(-\alpha)}{\sqrt{2-1}} & \cos(-\alpha)
\end{matrix}
\right ]
$. \\
A point $y'$ lies in the edge-rotated cone region of $y$ if and only if for some $\lambda$, $y' =  y + \lambda_1 A e_1 + \lambda_2 A e_2$ such that $\lambda_1, \lambda_2 \geq 0, \lambda \neq 0$. This is equivalent to: for some $\lambda$, $A^{-1}(y'-y) = \lambda_1 e_1 + \lambda e_2$ such that $\lambda_1, \lambda_2 \geq 0, \lambda \neq 0$. In short, \emph{ $y$ dominates $y'$ with respect to the edge-rotated cone order if and only if
the components of $A^{-1}(y'-y) $ are non-negative and at least one of them is strictly positive.}  
Thus, once the inverse matrix of $A$ is 
computed 
($A^{-1} =
c\cdot
\left [
\begin{matrix}
\cos (\alpha) & \sin(\alpha) \\
\sin(\alpha) & \cos(\alpha)
\end{matrix}
\right ]
$,\\
$c := \frac{1}{(\cos(\alpha))^2 - (\sin(\alpha))^2} $) 
, it can readily be determined whether $y'$ is in the dominating region of $y$. Moreover, in case the components are non-zero and have opposite signs, then the points are incomparable. In case the components are non-positive and at least one them negative, then $y'$ dominates $y$.

The approach can easily be applied to three or many objective problems. When the number of objectives is $m ~(m>2)$ and the rotation angle for each edge of the cone is $\alpha$, the ($m\times m$) matrix (\ref{matrix}) gives the coordinates of the unit point on rotated edges:
for each unit point on the edge of the standard Pareto cone, each 
column of the matrix gives its new coordinates after rotation. For example, in three-dimensional space, (1, 0, 0) is the unit point on one edge of the standard Pareto cone, then $(\cos{(-\alpha)}, \frac{\sin{(-\alpha)}}{\sqrt{2}}, \frac{\sin{(-\alpha)}}{\sqrt{2}} )$ are its new coordinates after the edge is rotated by an angle of $\alpha$ ($0 \leq \alpha < \frac{\pi}{4}$).
\begin{equation}
\left[
\begin{matrix}
 \cos{(-\alpha)}      & \frac{\sin{(-\alpha)}}{\sqrt{m-1}}      & \cdots & \frac{\sin{(-\alpha)}}{\sqrt{m-1}}      \\
 \frac{\sin{(-\alpha)}}{\sqrt{m-1}}      & \cos{(-\alpha)}      & \cdots & \frac{\sin{(-\alpha)}}{\sqrt{m-1}}      \\
 \vdots & \vdots & \ddots & \vdots \\
 \frac{\sin{(-\alpha)}}{\sqrt{m-1}}      & \frac{\sin{(-\alpha)}}{\sqrt{m-1}}      & \cdots & \cos{(-\alpha)}      \\
\end{matrix}
\right]
\label{matrix}
\end{equation}

When integrating the edge-rotated cone in MOEAs, the inverse matrix needs to be calculated only once. Therefore, almost no extra computing time is involved by Algorithm 1. For a similar cone construction, see \cite{emmerich2013}.

\vspace{-.2cm}
\section{Experimental Results and Discussion}
\label{sec:experiment}
\subsection{Experimental Design}
The proposed edge-rotated cone order can be integrated in all MOEAs using the Pareto order to select solutions. In this section, Algorithm 1 is combined in NSGA-II, DI-MOEA and NSGA-III to investigate the performance of the proposed approach when different rotation angles (i.e., from $3^\circ$ to $30^\circ$) have been applied. Four, six and eight objective DTLZ1, DTLZ2, DTLZ2\_convex problems have been chosen in the experiments. The optimal Pareto front of DTLZ1 lies on a linear hyperplane and the optimal Pareto front of DTLZ2 is concave. At the same time, to measure the performance on a convex problem, we transform DTLZ2 problem to DTLZ2\_convex problem with a convex Pareto front by simply decreasing all objective values by $3.5$. The other two benchmark problems include UF11 and UF13 \cite{zhang2008multiobjective}. UF11 is a rotated instance of the 5D DTLZ2 test problem, and UF13 is the 5D WFG1 test problem.

The population size is $100$ for all problems. We have taken 15 independent runs (with a different seed for each run but the same seed for each of the algorithms) of each algorithm on each problem. For each problem, the number of evaluations (NE) is the computing budget for running the algorithm and it is determined by $\max \{100000,10000\times D\}$, where $D$ is the number of decision variables. Two widely-used quality metrics, hypervolume (HV) \cite{while2006faster} and inverted generational distance (IGD) \cite{zitzler2003performance}, have been adopted to compare the performance of the algorithms. All experiments are implemented based on the MOEA Framework 2.12 (\url{http://www.moeaframework.org/}), which is a Java-based framework for multi-objective optimization. When calculating HV, the objective values of the reference point are 0.6 on DTLZ1, 1.1 on DTLZ2, 5 on DTLZ2\_convex, 2.2 on UF11 and 11 on UF13. The origin is used as the ideal point. When calculating the IGD value, the merged non-dominated solution sets from all runs are used as the reference sets of the DTLZ2\_convex problems and the reference sets of other problems are from the MOEA framework. 

\vspace{-0.2cm}
\subsection{Experimental Results}
Tables~\ref{dtlz2-concave} -~\ref{uf} show the mean hypervolume and IGD from 15 runs of DTLZ2 and UF problems when different edge-rotated cone orders are integrated in NSGA-II, DI-MOEA and NSGA-III. Tables for DTLZ1 and DTLZ2\_convex problems are in the appendix. The `` P\_cone'' column provides the results obtained by the original MOEAs. The ``$\frac{\pi}{6}(=30^{\circ})$'' column gives the results when each edge of the standard Pareto order cone has been rotated by 30$^{\circ}$ in the algorithm, similar remark for the other columns. The mean hypervolume and IGD values obtained by the original NSGA-II, DI-MOEA and NSGA-III have been used as the reference values to be compared with the results achieved by the algorithms involving the edge-rotated cone orders. For the algorithms combining the edge-rotated cone, the mean hypervolume and IGD values better than the values obtained by the original MOEAs have been highlighted in bold (i.e., a larger hypervolume value and lower IGD value); the largest respectively lowest value for each algorithm among them is printed in red. At the same time, the standard deviation of each algorithm is also given under each mean hypervolume and IGD. Tables for the DTLZ benchmark problems consist of four parts, namely four objective, six objective, eight objective with full budget, and eight objective with half budget. The behaviours of UF11 and UF13 with full budget and half budget are given in Table~\ref{uf}. Furthermore, the ranking of these algorithms has been calculated based on the mean hypervolume and shown in the appendix. 

\begin{table}[!htbp]
\centering
\caption{The mean Hypervolume (HV) and IGD on DTLZ2 (concave).}
\begin{tabular}{|c|c|c|c|c|c|c|c|c|}
\hline
\multicolumn{9}{|c|}{\bfseries Four objective (NE = 130000)}\\
\hline
Algorithms &Metrics & P\_cone & $\frac{\pi}{6}$(=30$^{\circ}$) & $\frac{\pi}{9}$(=20$^{\circ}$) & $\frac{\pi}{12}$(=15$^{\circ}$) & $\frac{\pi}{18}$(=10$^{\circ}$) &
$\frac{\pi}{30}$(=6$^{\circ}$) & $\frac{\pi}{60}$(=3$^{\circ}$) \\
\hline
\multirow{2}{*}{NSGA-II} & Mean HV & 0.5953 & 0.1971 & 0.5458 & \bfseries \color{red}0.6760 & \bfseries 0.6525 & \bfseries 0.6388 & \bfseries 0.6333  \\
\cline{2-9}
 & std & 0.0089 & 0.1182 & 0.0535 & 0.0041 & 0.0048 & 0.0080 & 0.0077 \\
\cline{1-9}

\multirow{2}{*}{DI-MOEA} & Mean HV & 0.6471 & 0.0913 & 0.5639 & \bfseries \color{red}0.6944 & \bfseries 0.6897 & \bfseries 0.6755 & \bfseries 0.6688\\
\cline{2-9}
 & std & 0.0094 & 0.0012 & 0.0406 & 0.0038 & 0.0026 & 0.0066 & 0.0039 \\
\cline{1-9}

\multirow{2}{*}{NSGA-III} & Mean HV & 0.6597 & 0.2508 & 0.5749 & \bfseries \color{red}0.6863 & \bfseries 0.6821 & \bfseries 0.6652 & 0.6592  \\
\cline{2-9}
 & std & 0.0054 & 0.1265 & 0.0362 & 0.0017 & 0.0040 & 0.0031 & 0.0066 \\
\hline
\hline

\multirow{2}{*}{NSGA-II} & Mean IGD & 0.1634 & 0.8352 & 0.4037 & 0.1867 & \bfseries \color{red}0.1492 & \bfseries 0.1536 & \bfseries 0.1542 \\
\cline{2-9}
& std & 0.0045 & 0.2290 & 0.0794 & 0.0056 & 0.0040 & 0.0055 & 0.0041 \\
\cline{1-9}

\multirow{2}{*}{DI-MOEA} & Mean IGD  & 0.1363 & 1.0405 & 0.3810 & 0.1731 &  \bfseries \color{red}0.1264 & \bfseries 0.1295 & \bfseries 0.1279 \\
\cline{2-9}
 & std &  0.0045 & 0.0183 & 0.0661 & 0.0049 & 0.0022 & 0.0061 & 0.0028 \\
\cline{1-9}

\multirow{2}{*}{NSGA-III} & Mean IGD & 0.1501 & 0.7553 & 0.3510 & 0.1749 & \bfseries \color{red}0.1361 & \bfseries 0.1477 & \bfseries 0.1490  \\
\cline{2-9}
 & std & 0.0046 & 0.2196 & 0.0705 & 0.0039 & 0.0034 & 0.0054 &0.0026 \\

\hline
\multicolumn{9}{|c|}{\bfseries Six objective (NE = 150000)}\\
\hline
\multirow{2}{*}{NSGA-II} & Mean HV & 0.1224 & 0.0000 & \bfseries 0.4304 & \bfseries \color{red}0.8156 & \bfseries 0.7608 & \bfseries 0.7284 & \bfseries 0.6490  \\
\cline{2-9}
 & std & 0.0701 & 0.0000 & 0.0254 & 0.0036 & 0.0067 & 0.0119 & 0.0221 \\
\cline{1-9}

\multirow{2}{*}{DI-MOEA} & Mean HV & 0.0000 & 0.0000 & \bfseries 0.4488 & \bfseries \color{red}0.8397 & \bfseries 0.8016 & \bfseries 0.7479 & \bfseries 0.6543\\
\cline{2-9}
 & std & 0.0000 & 0.0000 & 0.0126 & 0.0055 & 0.0055 & 0.0117 & 0.0347 \\
\cline{1-9}

\multirow{2}{*}{NSGA-III} & Mean HV & 0.8052 & 0.0000 & 0.4411 & \bfseries \color{red}0.8446 & \bfseries 0.8185 & \bfseries 0.8127 & \bfseries 0.8111  \\
\cline{2-9}
 & std & 0.0076 & 0.0000 & 0.0130 & 0.0048 & 0.0038 & 0.0056 & 0.0041 \\
\hline
\hline

\multirow{2}{*}{NSGA-II} & Mean IGD & 0.7278 & 2.5612 & 0.7003 & \bfseries 0.3447 & \bfseries \color{red}0.2856 & \bfseries 0.2887 & \bfseries 0.3137 \\
\cline{2-9}
& std & 0.0758 & 0.0090 & 0.0380 & 0.0119 & 0.0051 & 0.0046 & 0.0091 \\
\cline{1-9}

\multirow{2}{*}{DI-MOEA} & Mean IGD  & 1.9390 & 2.5824 & \bfseries 0.6961 & \bfseries 0.2913 & \bfseries \color{red}0.2774 & \bfseries 0.2898 & \bfseries 0.3335 \\
\cline{2-9}
 & std & 0.3246 & 0.0059 & 0.0285 & 0.0074 & 0.0026 & 0.0058 & 0.0172 \\
\cline{1-9}

\multirow{2}{*}{NSGA-III} & Mean IGD & 0.3125 & 2.5596 &  0.7260 & \bfseries  0.3073 & \bfseries \color{red}0.3061 & \bfseries 0.3092 & \bfseries 0.3095  \\
\cline{2-9}
 & std & 0.0105 & 0.0154 & 0.0283 & 0.0145 & 0.0071 & 0.0065 & 0.0080 \\

\hline
\multicolumn{9}{|c|}{\bfseries Eight objective (NE = 170000)}\\
\hline
\multirow{2}{*}{NSGA-II} & Mean HV & 0.0168 & 0.0000 & \bfseries 0.4947 & \bfseries \color{red}0.8850 & \bfseries 0.8193 & \bfseries 0.7068 & \bfseries 0.4062 \\
\cline{2-9}
 & std & 0.0355 & 0.0000 & 0.0576 & 0.0068 & 0.0068 & 0.0487 & 0.0754 \\
\cline{1-9}

\multirow{2}{*}{DI-MOEA} & Mean HV & 0.0000 & 0.0000 & \bfseries 0.4250 & \bfseries \color{red}0.9002 & \bfseries 0.8011 &\bfseries 0.4619 & \bfseries 0.0138\\
\cline{2-9}
 & std & 0.0000 & 0.0000 & 0.1260 & 0.0033 & 0.0196 & 0.1500 & 0.0516 \\
\cline{1-9}

\multirow{2}{*}{NSGA-III} & Mean HV & 0.8543 & 0.0000 & 0.3151 & \bfseries \color{red}0.9079 & \bfseries 0.8727 & \bfseries 0.8632 & 0.8522  \\
\cline{2-9}
 & std & 0.0121 & 0.0000 & 0.0643 & 0.0044 & 0.0074 & 0.0078 & 0.0138 \\
\hline
\hline

\multirow{2}{*}{NSGA-II} & Mean IGD & 1.2941 & 2.4798 & \bfseries 0.7887 & \bfseries 0.5247 & \bfseries \color{red}0.3955 & \bfseries  0.4332  & \bfseries 0.6433 \\
\cline{2-9}
& std & 0.1867 & 0.0422 & 0.0507 & 0.0210 & 0.0068 & 0.0201 & 0.0687 \\
\cline{1-9}

\multirow{2}{*}{DI-MOEA} & Mean IGD  & 2.4722 & 2.5704 & \bfseries  0.8728 &\bfseries  0.4483 & \bfseries \color{red}  0.4425 & \bfseries 0.6013 & \bfseries 2.3017  \\
\cline{2-9}
 & std &  0.0430 & 0.0129 & 0.1118 & 0.0054 & 0.0088 & 0.0682 & 0.4257 \\
\cline{1-9}

\multirow{2}{*}{NSGA-III} & Mean IGD & 0.4594 & 1.9278 & 0.9662 & 0.4936 & 0.4659 & 0.4638 & 0.4680\\
\cline{2-9}
 & std & 0.0105 & 0.1043 & 0.0491 & 0.0130 & 0.0099 & 0.0093 & 0.0175 \\

\hline
\multicolumn{9}{|c|}{\bfseries Eight objective - Half budget (NE = 85000)}\\
\hline
\multirow{2}{*}{NSGA-II} & Mean HV & 0.0001 & 0.0000 & \bfseries 0.4674 & \bfseries \color{red}0.8859 & \bfseries  0.8161 & \bfseries 0.7145 & \bfseries 0.4251 \\
\cline{2-9}
 & std & 0.0003 & 0.0000 & 0.0847 &  0.0047 & 0.0083 &0.0334 & 0.0851 \\
\cline{1-9}

\multirow{2}{*}{DI-MOEA} & Mean HV & 0.0000 & 0.0000 & \bfseries 0.4196 & \bfseries \color{red}0.9000 & \bfseries 0.8061 & \bfseries 0.5432 & \bfseries 0.0213\\
\cline{2-9}
 & std & 0.0000 & 0.0000 & 0.1254 & 0.0050 & 0.0207 & 0.0931 & 0.0606 \\
\cline{1-9}

\multirow{2}{*}{NSGA-III} & Mean HV & 0.8526 & 0.0000 & 0.3223 & \bfseries \color{red}0.9063 & \bfseries 0.8728 & \bfseries 0.8616 & \bfseries 0.8548  \\
\cline{2-9}
 & std & 0.0084 & 0.0000 & 0.0553 & 0.0048 & 0.0054 & 0.0085 & 0.0116 \\
\hline
\hline

\multirow{2}{*}{NSGA-II} & Mean IGD & 1.6856 &  2.4963 & \bfseries 0.8125 & \bfseries 0.5167 & \bfseries \color{red}0.3939 & \bfseries  0.4295  & \bfseries 0.6116 \\
\cline{2-9}
& std & 0.1949 & 0.0202 & 0.0763 & 0.0091 & 0.0060 & 0.0126 & 0.0869 \\
\cline{1-9}

\multirow{2}{*}{DI-MOEA} & Mean IGD  & 2.4858 &  2.5688 & \bfseries 0.8765 & \bfseries  0.4520 & \bfseries \color{red}0.4391 & \bfseries 0.5633 &\bfseries 2.0740  \\
\cline{2-9}
 & std & 0.0272 & 0.0276 & 0.1149 & 0.0073 & 0.0072 & 0.0403 & 0.5132 \\
\cline{1-9}

\multirow{2}{*}{NSGA-III} & Mean IGD & 0.4611 & 1.9307 & 0.9590 &  0.4923 & 0.4691 & 0.4630 & \bfseries \color{red}0.4597\\
\cline{2-9}
 & std &  0.0178 & 0.1646 & 0.0433 & 0.0127 & 0.0115 & 0.0101 & 0.0152 \\

\hline
\end{tabular}
\label{dtlz2-concave}
\end{table} 

\begin{table}[!htbp]
\centering
\caption{The mean Hypervolume (HV) and IGD on UF11 \& UF13.}
\begin{tabular}{|c|c|c|c|c|c|c|c|c|}
\hline
\multicolumn{9}{|c|}{\bfseries UF11 Five objective (NE = 300000)}\\
\hline
Algorithms &Metrics & P\_cone & $\frac{\pi}{6}$(=30$^{\circ}$) & $\frac{\pi}{9}$(=20$^{\circ}$) & $\frac{\pi}{12}$(=15$^{\circ}$) & $\frac{\pi}{18}$(=10$^{\circ}$) &
$\frac{\pi}{30}$(=6$^{\circ}$) & $\frac{\pi}{60}$(=3$^{\circ}$) \\
\hline
\multirow{2}{*}{NSGA-II} & Mean HV & 0.0000 & 0.0000 & \bfseries 0.0211 & \bfseries 0.0291 & \bfseries \color{red}0.0306 & \bfseries 0.0218 & \bfseries 0.0104  \\
\cline{2-9}
 & std & 0.0000 & 0.0000 & 0.0024 & 0.0058 & 0.0012 & 0.0011 & 0.0014 \\
\cline{1-9}

\multirow{2}{*}{DI-MOEA} & Mean HV & 0.0029 & 0.0000 & \bfseries 0.0191 & \bfseries \color{red}0.0336 & \bfseries 0.0256 & \bfseries 0.0188 & \bfseries 0.0138\\
\cline{2-9}
 & std &  0.0018 & 0.0000 & 0.0035 & 0.0008 & 0.0012 & 0.0015 & 0.0024 \\
\cline{1-9}

\multirow{2}{*}{NSGA-III} & Mean HV & 0.0147 &0.0000 & \bfseries  0.0266 & \bfseries \color{red} 0.0350 & \bfseries 0.0278 & \bfseries 0.0201 & \bfseries 0.0171  \\
\cline{2-9}
 & std & 0.0016 & 0.0000 & 0.0034 & 0.0017 & 0.0016 & 0.0014 & 0.0015 \\
\hline
\hline

\multirow{2}{*}{NSGA-II} & Mean IGD & 1.5208 & 14.6626 & \bfseries 0.3890 & \bfseries 0.2990 & \bfseries \color{red}0.2685 & \bfseries 0.3119 & \bfseries 0.4531 \\
\cline{2-9}
& std & 0.2173 & 0.2878 & 0.0368 & 0.0374 & 0.0171 & 0.0241 & 0.0289 \\
\cline{1-9}

\multirow{2}{*}{DI-MOEA} & Mean IGD  & 0.7304 & 15.1690 & \bfseries 0.6152 & \bfseries \color{red}0.2807 &  \bfseries 0.3339 & \bfseries 0.3946 & \bfseries 0.4621 \\
\cline{2-9}
 & std & 0.0944 & 0.2054 & 0.1997 & 0.0210 & 0.0228 & 0.0352 & 0.0545 \\
\cline{1-9}

\multirow{2}{*}{NSGA-III} & Mean IGD & 0.4517 & 15.0785 & \bfseries 0.4190 & \bfseries \color{red}0.2795 & \bfseries 0.3188 & \bfseries 0.3848 & \bfseries 0.4166  \\
\cline{2-9}
 & std & 0.0388 & 0.2105 & 0.0697 & 0.0247 & 0.0235 & 0.0324 & 0.0183 \\

\hline
\multicolumn{9}{|c|}{\bfseries UF11 Five objective - Half budget (NE = 150000)}\\
\hline
\multirow{2}{*}{NSGA-II} & Mean HV & 0.0000 & 0.0000 & \bfseries 0.0205 & \bfseries 0.0269 & \bfseries \color{red}0.0288 & \bfseries 0.0201 & \bfseries 0.0082  \\
\cline{2-9}
 & std & 0.0000 & 0.0000 & 0.0025 &  0.0055 & 0.0014 & 0.0016 & 0.0017 \\
\cline{1-9}

\multirow{2}{*}{DI-MOEA} & Mean HV & 0.0012 & 0.0000 & \bfseries 0.0237 & \bfseries \color{red}0.0316 & \bfseries 0.0244 & \bfseries 0.0185 & \bfseries 0.0126\\
\cline{2-9}
 & std & 0.0011 & 0.0000 & 0.0030 &  0.0020 & 0.0010 & 0.0014 & 0.0017 \\
\cline{1-9}

\multirow{2}{*}{NSGA-III} & Mean HV & 0.0148 & 0.0000 & \bfseries 0.0268 &\bfseries \color{red}0.0342 & \bfseries 0.0270 & \bfseries 0.0199 & \bfseries 0.0170 \\
\cline{2-9}
 & std & 0.0020 & 0.0000 & 0.0029 & 0.0013 & 0.0018 &  0.0016 & 0.0010 \\
\hline
\hline

\multirow{2}{*}{NSGA-II} & Mean IGD &  1.7202 & 14.7243 & \bfseries 0.3951 & \bfseries 0.3031 & \bfseries \color{red}0.2731 & \bfseries  0.3208 & \bfseries 0.4846 \\
\cline{2-9}
& std &  0.2541 & 0.1769 & 0.0392 & 0.0343 & 0.0164 & 0.0289 & 0.0312 \\
\cline{1-9}

\multirow{2}{*}{DI-MOEA} & Mean IGD  & 0.8730 & 15.1172 & \bfseries 0.4910 & \bfseries \color{red}0.2939 & \bfseries 0.3418 & \bfseries 0.4061 & \bfseries 0.4831 \\
\cline{2-9}
 & std & 0.1485 & 0.2099 & 0.0619 & 0.0269 & 0.0244 & 0.0329 & 0.0439 \\
\cline{1-9}

\multirow{2}{*}{NSGA-III} & Mean IGD & 0.4606 & 15.0148 & \bfseries 0.3897 & \bfseries \color{red}0.2752 & \bfseries 0.3204 & \bfseries 0.4009 & \bfseries 0.4314  \\
\cline{2-9}
 & std & 0.0433 & 0.1881 & 0.0615 & 0.0186 & 0.0265 & 0.0393 & 0.0335 \\

\hline
\multicolumn{9}{|c|}{\bfseries UF13 Five objective (NE = 300000)}\\
\hline
\multirow{2}{*}{NSGA-II} & Mean HV & 0.6937 & 0.5041 & \bfseries 0.7410 & \bfseries \color{red}0.7424 & \bfseries 0.7177 & \bfseries 0.7065 & \bfseries 0.6994 \\
\cline{2-9}
 & std & 0.0079 & 0.1742 & 0.0096  &0.0070 & 0.0091 & 0.0084 & 0.0084 \\
\cline{1-9}
\multirow{2}{*}{DI-MOEA} & Mean HV & 0.6611 & 0.4625 &  \bfseries \color{red}0.7343 & \bfseries 0.7152 & 0.6590 & 0.6567 & 0.6589\\
\cline{2-9}
 & std & 0.0063 & 0.1580 & 0.0064 & 0.0119 & 0.0073 & 0.0067 & 0.0071 \\
\cline{1-9}

\multirow{2}{*}{NSGA-III} & Mean HV & 0.6498 & 0.4523 & \bfseries 0.7164 & \bfseries \color{red}0.7226 & \bfseries 0.7023 & \bfseries 0.6703 & \bfseries 0.6532  \\
\cline{2-9}
 & std &0.0130 & 0.1017 & 0.0048 & 0.0108 & 0.0085 &  0.0106 & 0.0077 \\
\hline
\hline

\multirow{2}{*}{NSGA-II} & Mean IGD & 1.4761 & \bfseries 1.3108 & \bfseries 1.4316 & \bfseries \color{red}1.3805 & \bfseries 1.4656 & \bfseries 1.4391  & \bfseries 1.4181 \\
\cline{2-9}
& std & 0.1315 & 0.2267 & 0.0565 & 0.0857 & 0.0664 & 0.1572 & 0.1029 \\
\cline{1-9}

\multirow{2}{*}{DI-MOEA} & Mean IGD  & 1.5448 & \bfseries \color{red}1.5031 & \bfseries 1.5151 & 1.5481 & 1.7512 & 1.6351 & 1.5934  \\
\cline{2-9}
 & std & 0.0473 & 0.4180 & 0.0533 & 0.0646 & 0.0384 & 0.0667 &  0.0399 \\
\cline{1-9}

\multirow{2}{*}{NSGA-III} & Mean IGD & 1.8698 &\bfseries 1.6030 & \bfseries 1.6324 & \bfseries \color{red} 1.5813 & \bfseries 1.6675 & \bfseries 1.7950 & \bfseries 1.8527\\
\cline{2-9}
 & std & 0.1842 & 0.1835 & 0.0285 & 0.0658 & 0.0969 &  0.1457 &  0.1245 \\
\hline

\multicolumn{9}{|c|}{\bfseries UF13 Five objective - Half budget (NE = 150000)}\\
\hline
\multirow{2}{*}{NSGA-II} & Mean HV & 0.6687 & 0.5016 & \bfseries \color{red}0.7259 & \bfseries 0.7170 & \bfseries 0.6915 & \bfseries 0.6831 & \bfseries 0.6738 \\
\cline{2-9}
 & std & 0.0041 & 0.1749 & 0.0092 & 0.0058 & 0.0042 & 0.0047 & 0.0057 \\
\cline{1-9}
\multirow{2}{*}{DI-MOEA} & Mean HV & 0.6457 & 0.3427 & \bfseries \color{red}0.7254 & \bfseries 0.7002 & \bfseries 0.6513 & \bfseries 0.6481 & \bfseries 0.6497\\
\cline{2-9}
 & std & 0.0045 & 0.2041 & 0.0044 & 0.0133 & 0.0056 & 0.0053 & 0.0057 \\
\cline{1-9}

\multirow{2}{*}{NSGA-III} & Mean HV & 0.6432 & 0.4702 & \bfseries \color{red}0.7073 & \bfseries 0.7045 & \bfseries 0.6770 & \bfseries 0.6579 & 0.6417  \\
\cline{2-9}
 & std & 0.0086 & 0.0996 & 0.0074 & 0.0076 & 0.0103 & 0.0071 & 0.0056 \\
\hline
\hline
\multirow{2}{*}{NSGA-II} & Mean IGD & 1.5720 & 1.3736 & \bfseries 1.5455 & \bfseries \color{red}1.5074 & 1.5968 & 1.5746  & \bfseries 1.5262 \\
\cline{2-9}
& std & 0.0946 & 0.1703 & 0.0638 & 0.0649 & 0.0786 & 0.1135 & 0.0860 \\
\cline{1-9}

\multirow{2}{*}{DI-MOEA} & Mean IGD  & 1.6609 & \bfseries \color{red}1.5321 & \bfseries 1.5939 & \bfseries 1.6311 & 1.8048 & 1.7286& 1.6403  \\
\cline{2-9}
 & std & 0.0557 & 0.3781 & 0.0268 & 0.0781 & 0.0509 & 0.0794 & 0.0613 \\
\cline{1-9}

\multirow{2}{*}{NSGA-III} & Mean IGD &  1.8931 & 1.7553 & \bfseries \color{red}1.6824 &  \bfseries 1.6832 & \bfseries 1.8163 & 1.8976 &  1.9725\\
\cline{2-9}
 & std & 0.1238 & 0.2361 & 0.0456 & 0.0376 & 0.0924 & 0.1200 & 0.0562 \\
\hline
\end{tabular}
\label{uf}
\end{table}

We can draw the following conclusions from the data in these tables. 
\vspace{-0.1cm}
\begin{enumerate}
\item The algorithms do not work well when a large rotation angle is adopted (e.g., $30^{\circ}$); only the mean rank of the algorithm involving the cone with a $30^{\circ}$ rotation angle is worse than the mean rank of the original MOEA. 

\item The algorithms show similar performance to the original MOEAs when the rotation angle is very small (e.g., $3^{\circ}$).

\item When an intermediate rotation angle is adopted, the performance of the algorithms (both hypervolume and IGD values) shows a significant improvement except for a few cases which display values close to the original MOEAs. 

\item Although it differs depending on the specific problems, the best performance is usually obtained when the rotation angle is $15^{\circ}$. Also, the mean rank of the algorithm involving the cone with a $15^{\circ}$ rotation angle is the best and a $10^{\circ}$ rotation angle is the second best.

\item It can be seen that the edge-rotated cone can improve the performance of all three adopted MOEAs (i.e., NSGA-II, DI-MOEA and NSGA-III) in most cases when an intermediate rotation angle is used. Even though NSGA-III is assumed to be powerful enough to handle these benchmark problems, its performance can still be improved by the edge-rotated cone approach. 

\item The edge-rotated cone can benefit MOEAs even more with the increase of the number of objectives. For example, when a $15^{\circ}$ rotation angle is applied on the DTLZ2 (concave) four objective problem, the hypervolume of NSGA-II is improved from 0.5953 to 0.6760; for the six objective problem, the hypervolume is improved from 0.1224 to 0.8156; and for the  eight objective problem, the hypervolume is improved from 0.0168 to 0.8850. 

\item The edge-rotated cone can benefit the algorithm with a small computing budget more than the algorithm with a large budget. For example, when using half of the computing budget on UF13 five objective problem and the rotation angle is set to $20^{\circ}$, the hypervolume values of the Pareto fronts from NSGA-II, DI-MOEA and NSGA-III can be improved to 0.7259, 0.7254, 0.7073, which are already larger than the hypervolume values obtained by the original MOEAs with full budget, namly 0.6937, 0.6611 and 0.6497.

\item Even though we did not show the median values of the hypervolume and IGD values in the tables, they show similar values as the mean values. At the same time, the standard deviations show a stable behavior of the edge-rotated cone order when it is integrated in MOEAs.

\end{enumerate}

\section{Conclusions and Further Work}
\label{sec:conclusion}
In this paper, we enhance the standard Pareto dominance relationship from the geometric perspective. By rotating the edges of the standard Pareto order cone, the incomparable solutions can be ranked into different layers, hence, the selection pressure toward the Pareto front can be strengthened and the convergence of the algorithm can be accelerated. To avoid neglecting the diversity, the edge-rotated cone order is designed to work together with the standard Pareto order in our algorithm. After testing various angles on different many-objective optimization problems, we show the ability of improving the performance of original MOEAs by the edge-rotated cone and suggest that the rotation angle of $15^{\circ}$ can be adopted in the absence of specific experiments or knowledge of the application domain. Our method of implementing the integration of the edge-rotated cone barely needs more computing time compared to the original MOEAs, moreover, with a small computing budget, it can promote the performance of the algorithm to the effect of using a large budget without using the edge-rotated cone orders.

Our implementation of enhancing the Pareto dominance is straightforward and effective, we think it is a good direction to improve any MOEA using the Pareto dominance to select solutions. In future, the mechanism that relates the properties of the problem with the rotation angle should be researched. Another interesting direction of future work could be to investigate and compare different schemes of alternating between the cone orders in order to promote diversity and convergence. For instance, it could be investigated whether using acute cones can be of benefit to promote diversity even more, or to use again the Pareto cone in the final stage of the evolution to make sure that no solutions are excluded from the Pareto front which might happen when using the edge-rotated cone.

%
%
%
%

\appendix

\section{Appendix: Tables for Section 4}
\label{appendix:a}

\begin{table}[!htbp]
\centering
\caption{The mean Hypervolume (HV) and IGD on DTLZ1.}
\begin{tabular}{|c|c|c|c|c|c|c|c|c|}
\hline
\multicolumn{9}{|c|}{\bfseries Four objective (NE = 100000)}\\
\hline
Algorithms &Metrics & P\_cone & $\frac{\pi}{6}$(=30$^{\circ}$) & $\frac{\pi}{9}$(=20$^{\circ}$) & $\frac{\pi}{12}$(=15$^{\circ}$) & $\frac{\pi}{18}$(=10$^{\circ}$) &
$\frac{\pi}{30}$(=6$^{\circ}$) & $\frac{\pi}{60}$(=3$^{\circ}$) \\
\hline
\multirow{2}{*}{NSGA-II} & Mean HV & 0.5811 & \bfseries 0.7735 & \bfseries \color{red}0.9405 & \bfseries 0.9403 & \bfseries 0.9400 & \bfseries 0.9393 & \bfseries 0.9398  \\
\cline{2-9}
 & std & 0.3347 & 0.1918 & 0.0024 & 0.0021 & 0.0026 & 0.0021 & 0.0018 \\
\cline{1-9}

\multirow{2}{*}{DI-MOEA} & Mean HV & 0.4842 & 0.0000 & \bfseries 0.9535 & \bfseries 0.9537 & \bfseries 0.9521 & \bfseries \color{red}0.9538 & \bfseries 0.9533\\
\cline{2-9}
 & std & 0.4250 & 0.0000 & 0.0010 & 0.0009 & 0.0056 & 0.0005 & 0.0009 \\
\cline{1-9}

\multirow{2}{*}{NSGA-III} & Mean HV & 0.9447 & 0.6399 & \bfseries 0.9448 & 0.9444 & \bfseries \color{red}0.9458 & \bfseries 0.9453 & \bfseries 0.9452  \\
\cline{2-9}
 & std & 0.0024 & 0.2561 & 0.0020 & 0.0024 & 0.0020 &  0.0018 & 0.0028 \\
\hline
\hline

\multirow{2}{*}{NSGA-II} & Mean IGD & 0.9725 & \bfseries 0.3083 & \bfseries 0.1537 & \bfseries 0.1550 & \bfseries 0.1553 & \bfseries \color{red}0.1491 & \bfseries 0.1511 \\
\cline{2-9}
& std & 0.0044 & 0.0444 & 0.0036 & 0.0026 & 0.0037 & 0.0046 & 0.0034 \\
\cline{1-9}

\multirow{2}{*}{DI-MOEA} & Mean IGD  & 1.2772 & 763.0901 & \bfseries \color{red}0.1287 & \bfseries \color{red}0.1287 &  \bfseries 0.1329 & \bfseries 0.1303 & \bfseries 0.1311  \\
\cline{2-9}
 & std & 1.4694 & 9.6585 & 0.0021 & 0.0026 & 0.0128 & 0.0019 & 0.0027 \\
\cline{1-9}

\multirow{2}{*}{NSGA-III} & Mean IGD & 0.1300 & 0.4122 & \bfseries 0.1297 & \bfseries \color{red}0.1295 & 0.1315 & \bfseries 0.1298 & 0.1313  \\
\cline{2-9}
 & std &  0.0024 & 0.2506 & 0.0038 & 0.0031 & 0.0029 & 0.0027 &0.0032 \\

\hline
\multicolumn{9}{|c|}{\bfseries Six objective (NE = 100000)}\\
\hline
\multirow{2}{*}{NSGA-II} & Mean HV & 0.0000 & 0.0000 & \bfseries \color{red}0.9857 & \bfseries 0.9851 & \bfseries 0.9844 & \bfseries 0.9808 & \bfseries 0.8922  \\
\cline{2-9}
 & std & 0.0000 & 0.0000 & 0.0007 & 0.0007 & 0.0010 & 0.0023 & 0.2001 \\
\cline{1-9}

\multirow{2}{*}{DI-MOEA} & Mean HV & 0.0000 & 0.0000 & \bfseries \color{red}0.9911 & \bfseries \color{red}0.9911 & \bfseries 0.9906 & \bfseries 0.9885 & \bfseries 0.9728\\
\cline{2-9}
 & std & 0.0000 & 0.0000 & 0.0002 & 0.0003 & 0.0002 & 0.0019 & 0.0084 \\
\cline{1-9}

\multirow{2}{*}{NSGA-III} & Mean HV & 0.9880 & 0.0000 & \bfseries \color{red}0.9887 & \bfseries 0.9885 & \bfseries 0.9883 & \bfseries 0.9881 & \bfseries 0.9883  \\
\cline{2-9}
 & std & 0.0009 & 0.0000 & 0.0005 & 0.0006 & 0.0005 & 0.0008 & 0.0006 \\
\hline
\hline

\multirow{2}{*}{NSGA-II} & Mean IGD & 75.4078 & 744.6850 & \bfseries 0.3041 & \bfseries \color{red}0.3026 & \bfseries 0.3079 & \bfseries 0.3256 & \bfseries 0.4541 \\
\cline{2-9}
& std & 41.1790 & 49.7971 & 0.0169 & 0.0136 & 0.0183 & 0.0159 & 0.2359 \\
\cline{1-9}

\multirow{2}{*}{DI-MOEA} & Mean IGD  & 349.0537 &  769.4755 & \bfseries \color{red}0.3086 & \bfseries 0.3102 & \bfseries 0.3151 & \bfseries 0.3196 & \bfseries 0.3791 \\
\cline{2-9}
 & std & 76.0015 & 37.4396 & 0.0050 & 0.0043 & 0.0064 & 0.0104 & 0.0291 \\
\cline{1-9}

\multirow{2}{*}{NSGA-III} & Mean IGD & 0.2990 & 770.0300 & \bfseries \color{red}0.2935 & 0.3007 & 0.3020 & 0.3015 & 0.3020  \\
\cline{2-9}
 & std & 0.0101 & 40.8585 & 0.0050 & 0.0085 & 0.0095 & 0.0085 & 0.0092 \\

\hline
\multicolumn{9}{|c|}{\bfseries Eight objective (NE = 120000)}\\
\hline
\multirow{2}{*}{NSGA-II} & Mean HV & 0.0000 & 0.0000 & \bfseries \color{red}0.9957 & \bfseries 0.9956 & \bfseries 0.9937 & \bfseries 0.9422 & \bfseries 0.7397 \\
\cline{2-9}
 & std & 0.0000 & 0.0000 & 0.0003 & 0.0004 & 0.0005 & 0.1638 & 0.3584 \\
\cline{1-9}
\multirow{2}{*}{DI-MOEA} & Mean HV & 0.0000 & 0.0000 & \bfseries \color{red}0.9976 & \bfseries \color{red}0.9976 & \bfseries 0.9965 & \bfseries 0.8700 & \bfseries 0.2850\\
\cline{2-9}
 & std & 0.0000 & 0.0000 & 0.0001 & 0.0002 & 0.0007 & 0.2892 & 0.3758 \\
\cline{1-9}

\multirow{2}{*}{NSGA-III} & Mean HV & 0.9877 & 0.0000 & 0.9855 & 0.9858 & 0.9853 & 0.9865 & 0.9854  \\
\cline{2-9}
 & std & 0.0025 & 0.0000 & 0.0027 & 0.0038 & 0.0032 & 0.0042 & 0.0025 \\
\hline
\hline

\multirow{2}{*}{NSGA-II} & Mean IGD & 128.0384 & 721.0803 & \bfseries 0.4286 & \bfseries \color{red}0.4272 & \bfseries 0.4452 & \bfseries 0.5575  & \bfseries 0.8845 \\
\cline{2-9}
& std & 56.8022 & 57.7441 & 0.0199 & 0.0148 & 0.0232 & 0.2231 & 0.4798 \\
\cline{1-9}

\multirow{2}{*}{DI-MOEA} & Mean IGD  & 517.2231 & 758.8918 & \bfseries \color{red}0.4843 & \bfseries 0.4866 & \bfseries 0.5043 & \bfseries 0.8457 & \bfseries 3.3619  \\
\cline{2-9}
 & std & 108.7324 & 142.8642 & 0.0068 & 0.0056 & 0.0106 & 0.6234 & 3.5900 \\
\cline{1-9}

\multirow{2}{*}{NSGA-III} & Mean IGD & 0.3599 & 418.6033 & \bfseries \color{red}0.3461 &  \bfseries 0.3567 & \bfseries 0.3565 & \bfseries 0.3557 & \bfseries 0.3594\\
\cline{2-9}
 & std & 0.0113 & 43.8714 & 0.0106 & 0.0106 & 0.0123 & 0.0192 & 0.0096 \\
\hline

\multicolumn{9}{|c|}{\bfseries Eight objective - Half budget (NE = 60000)}\\
\hline
\multirow{2}{*}{NSGA-II} & Mean HV & 0.0000 & 0.0000 & \bfseries \color{red}0.9954 & \bfseries 0.9944 & \bfseries 0.7331 & \bfseries 0.2971 & \bfseries 0.2048 \\
\cline{2-9}
 & std & 0.0000 & 0.0000 & 0.0006 & 0.0012 & 0.3764 & 0.3660 & 0.3120 \\
\cline{1-9}
\multirow{2}{*}{DI-MOEA} & Mean HV & 0.0000 & 0.0000 & \bfseries 0.9634 & \bfseries \color{red}0.9972 & \bfseries 0.9861 & \bfseries 0.7335 & \bfseries 0.0745\\
\cline{2-9}
 & std & 0.0000 & 0.0000 & 0.0863 & 0.0003 & 0.0341 & 0.3592 & 0.1555 \\
\cline{1-9}

\multirow{2}{*}{NSGA-III} & Mean HV & 0.9813 & 0.0000 & \bfseries \color{red}0.9855 & \bfseries 0.9842 & \bfseries 0.9849 & \bfseries 0.9856 & \bfseries 0.9863  \\
\cline{2-9}
 & std & 0.0138 & 0.0000 & 0.0027 & 0.0033 & 0.0033 & 0.0034 & 0.0027 \\
\hline
\hline
\multirow{2}{*}{NSGA-II} & Mean IGD & 170.7728 & 681.8762 & \bfseries \color{red}0.4248 & \bfseries \color{red}0.4248 & \bfseries 1.0092 & \bfseries 2.3994  & \bfseries 3.2542 \\
\cline{2-9}
& std & 92.0427 & 64.1913 & 0.0204 & 0.0151 & 0.8559 & 2.0576 & 3.1045 \\
\cline{1-9}

\multirow{2}{*}{DI-MOEA} & Mean IGD  & 592.0768 & 747.5064 & \bfseries 0.5889 & \bfseries \color{red}0.4881 & \bfseries  0.5406 & \bfseries 1.2579 & \bfseries 4.0558  \\
\cline{2-9}
 & std & 93.9853 & 94.6608 & 0.2782 & 0.0087 & 0.0983 & 1.1140 & 2.8857 \\
\cline{1-9}

\multirow{2}{*}{NSGA-III} & Mean IGD &  0.3777 & 405.6668 & \bfseries \color{red}0.3509 &  \bfseries 0.3576 & \bfseries 0.3635 & \bfseries 0.3663 & \bfseries 0.3627\\
\cline{2-9}
 & std & 0.0588 & 48.9879 & 0.0192 & 0.0107 & 0.0124 & 0.0216 & 0.0169 \\
\hline
\end{tabular}
\label{dtlz1}
\end{table} 

\begin{table}[!htbp]
\centering
\caption{The mean Hypervolume (HV) and IGD on DTLZ2\_convex.}
\begin{tabular}{|c|c|c|c|c|c|c|c|c|}
\hline
\multicolumn{9}{|c|}{\bfseries Four objective (NE = 130000)}\\
\hline
Algorithms &Metrics & P\_cone & $\frac{\pi}{6}$(=30$^{\circ}$) & $\frac{\pi}{9}$(=20$^{\circ}$) & $\frac{\pi}{12}$(=15$^{\circ}$) & $\frac{\pi}{18}$(=10$^{\circ}$) &
$\frac{\pi}{30}$(=6$^{\circ}$) & $\frac{\pi}{60}$(=3$^{\circ}$) \\
\hline
\multirow{2}{*}{NSGA-II} & Mean HV & 0.4433 & 0.2126 & 0.4302 & \bfseries \color{red}0.4613 & \bfseries 0.4577 & \bfseries 0.4514 & \bfseries 0.4502  \\
\cline{2-9}
 & std & 0.0046 & 0.0286 & 0.0025 & 0.0019 & 0.0027 & 0.0037 & 0.0036 \\
\cline{1-9}

\multirow{2}{*}{DI-MOEA} & Mean HV & 0.4643 & 0.0427 & 0.4308 & \bfseries 0.4673 & \bfseries \color{red}0.4730 & \bfseries 0.4688 & \bfseries 0.4678\\
\cline{2-9}
 & std & 0.0071 & 0.0048 & 0.0039 & 0.0051 & 0.0017 & 0.0025 & 0.0019 \\
\cline{1-9}

\multirow{2}{*}{NSGA-III} & Mean HV & 0.4419 & 0.1978 & 0.4182 & \bfseries 0.4501 & \bfseries \color{red}0.4552 & \bfseries 0.4499 & \bfseries 0.4470  \\
\cline{2-9}
 & std & 0.0078 & 0.0329 & 0.0037 & 0.0036 & 0.0025 &  0.0053 & 0.0036 \\
\hline
\hline

\multirow{2}{*}{NSGA-II} & Mean IGD & 0.1484 & 0.5018 & 0.2137 & 0.1512 & \bfseries \color{red}0.1454 & \bfseries 0.1466 & \bfseries 0.1458 \\
\cline{2-9}
& std & 0.0044 & 0.0444 & 0.0036 & 0.0026 & 0.0037 & 0.0046 & 0.0034 \\
\cline{1-9}

\multirow{2}{*}{DI-MOEA} & Mean IGD  & 0.1284 & 0.7288 & 0.2108 & 0.1426 &  \bfseries \color{red}0.1238 & \bfseries 0.1252 & \bfseries 0.1255  \\
\cline{2-9}
 & std & 0.0093 & 0.0154 & 0.0055 & 0.0074 & 0.0017 & 0.0034 & 0.0026 \\
\cline{1-9}

\multirow{2}{*}{NSGA-III} & Mean IGD & 0.1471 & 0.5242 & 0.2295 &  0.1660 & \bfseries \color{red}0.1424 & \bfseries 0.1439 & \bfseries 0.1424  \\
\cline{2-9}
 & std & 0.0094 & 0.0511 & 0.0052 & 0.0052 & 0.0031 & 0.0067 &0.0043 \\

\hline
\multicolumn{9}{|c|}{\bfseries Six objective (NE = 150000)}\\
\hline
\multirow{2}{*}{NSGA-II} & Mean HV & 0.1299 & 0.0223 & \bfseries 0.1304 & \bfseries \color{red}0.1471 & \bfseries 0.1376 & \bfseries 0.1348 & \bfseries 0.1325  \\
\cline{2-9}
 & std & 0.0029 & 0.0042 & 0.0016 & 0.0017 & 0.0018 & 0.0027 & 0.0023 \\
\cline{1-9}

\multirow{2}{*}{DI-MOEA} & Mean HV & 0.1343 & 0.0133 & 0.1280 & \bfseries \color{red}0.1525 & \bfseries 0.1408 & \bfseries 0.1376 & \bfseries 0.1365\\
\cline{2-9}
 & std & 0.0018 & 0.0009 & 0.0019 & 0.0011 & 0.0014 & 0.0017 & 0.0020 \\
\cline{1-9}

\multirow{2}{*}{NSGA-III} & Mean HV & 0.0993 & 0.0072 & \bfseries 0.1109 & \bfseries \color{red}0.1386 & \bfseries 0.1234 & \bfseries 0.1116 & \bfseries 0.1045  \\
\cline{2-9}
 & std & 0.0078 & 0.0010 & 0.0026 & 0.0027 & 0.0045 & 0.0061 & 0.0072 \\
\hline
\hline

\multirow{2}{*}{NSGA-II} & Mean IGD & 0.2713 & 0.5058 & 0.4106 & 0.2789 & \bfseries \color{red}0.2655 & \bfseries 0.2686 & \bfseries 0.2698 \\
\cline{2-9}
& std & 0.0047 & 0.0282 & 0.0043 & 0.0049 & 0.0046 & 0.0053 & 0.0054 \\
\cline{1-9}

\multirow{2}{*}{DI-MOEA} & Mean IGD  & 0.2571 &  0.6012 & 0.4149 & 0.2657 & \bfseries \color{red}0.2513 & \bfseries 0.2530 & \bfseries 0.2557 \\
\cline{2-9}
 & std & 0.0030 & 0.0044 & 0.0050 & 0.0054 & 0.0038 & 0.0029 & 0.0028 \\
\cline{1-9}

\multirow{2}{*}{NSGA-III} & Mean IGD & 0.2911 & 0.7106 &  0.4557 &  0.3039 & \bfseries \color{red}0.2677 & \bfseries 0.2764 & \bfseries 0.2869  \\
\cline{2-9}
 & std & 0.0093 & 0.0245 & 0.0074 & 0.0110 & 0.0106 & 0.0070 & 0.0073 \\

\hline
\multicolumn{9}{|c|}{\bfseries Eight objective (NE = 170000)}\\
\hline
\multirow{2}{*}{NSGA-II} & Mean HV & 0.0276 & 0.0155 & 0.0187 & \bfseries \color{red}0.0355 & \bfseries 0.0298 & \bfseries 0.0292 & \bfseries 0.0283 \\
\cline{2-9}
 & std & 0.0010 & 0.0013 & 0.0029 & 0.0005 & 0.0011 & 0.0007 & 0.0008 \\
\cline{1-9}

\multirow{2}{*}{DI-MOEA} & Mean HV & 0.0264 & 0.0213 & 0.0151 & \bfseries \color{red}0.0357 & \bfseries 0.0280 & \bfseries 0.0269 & \bfseries 0.0267\\
\cline{2-9}
 & std & 0.0008 & 0.0007 & 0.0004 & 0.0005 & 0.0006 & 0.0007 & 0.0009 \\
\cline{1-9}

\multirow{2}{*}{NSGA-III} & Mean HV & 0.0210 & 0.0014 & 0.0127 & \bfseries \color{red}0.0256 & \bfseries 0.0219 & \bfseries 0.0211 & 0.0206  \\
\cline{2-9}
 & std & 0.0010 & 0.0013 & 0.0005 & 0.0009 & 0.0015 & 0.0010 & 0.0014 \\
\hline
\hline

\multirow{2}{*}{NSGA-II} & Mean IGD & 0.3649 & 0.4218 & 0.5285 & \bfseries 0.3607 & \bfseries \color{red}0.3548 & \bfseries 0.3573  & \bfseries 0.3607 \\
\cline{2-9}
& std & 0.0087 & 0.0083 & 0.0236 & 0.0040 & 0.0061 & 0.0086 & 0.0067 \\
\cline{1-9}

\multirow{2}{*}{DI-MOEA} & Mean IGD  & 0.3816 & 0.3946 & 0.5611 & \bfseries \color{red}0.3597 & \bfseries 0.3736 & \bfseries 0.3788 & \bfseries 0.3803  \\
\cline{2-9}
 & std & 0.0036 & 0.0047 & 0.0029 & 0.0048 & 0.0044 & 0.0057 & 0.0050 \\
\cline{1-9}

\multirow{2}{*}{NSGA-III} & Mean IGD & 0.4197 & 0.7074 & 0.5811 &  \bfseries 0.4178 & \bfseries \color{red}0.4176 & 0.4198 & 0.4211\\
\cline{2-9}
 & std & 0.0094 & 0.0272 & 0.0037 & 0.0073 & 0.0136 & 0.0095 & 0.0120 \\

\hline
\multicolumn{9}{|c|}{\bfseries Eight objective - Half budget (NE = 85000)}\\
\hline
\multirow{2}{*}{NSGA-II} & Mean HV & 0.0282 & 0.0152 & 0.0187 & \bfseries \color{red}0.0356 & \bfseries 0.0304 & \bfseries 0.0293 & \bfseries 0.0286 \\
\cline{2-9}
 & std & 0.0007 & 0.0012 & 0.0025 & 0.0006 & 0.0006 & 0.0007 & 0.0010 \\
\cline{1-9}

\multirow{2}{*}{DI-MOEA} & Mean HV & 0.0263 & 0.0217 & 0.0150 & \bfseries \color{red}0.0359 & \bfseries 0.0276 & \bfseries 0.0268 & \bfseries 0.0266\\
\cline{2-9}
 & std & 0.0010 & 0.0008 & 0.0005 & 0.0006 & 0.0006 & 0.0007 & 0.0008 \\
\cline{1-9}

\multirow{2}{*}{NSGA-III} & Mean HV & 0.0202 & 0.0012 & 0.0126 & \bfseries \color{red}0.0249 & \bfseries 0.0213 & \bfseries 0.0207 & 0.0200  \\
\cline{2-9}
 & std & 0.0013 & 0.0009 & 0.0005 & 0.0009 & 0.0008 & 0.0010 & 0.0014 \\
\hline
\hline

\multirow{2}{*}{NSGA-II} & Mean IGD & 0.3649 & 0.4218 & 0.5285 & \bfseries 0.3607 & \bfseries \color{red}0.3548 & \bfseries 0.3573  & \bfseries 0.3607 \\
\cline{2-9}
& std & 0.0087 & 0.0083 & 0.0236 & 0.0040 & 0.0061 & 0.0086 & 0.0067 \\
\cline{1-9}

\multirow{2}{*}{DI-MOEA} & Mean IGD  & 0.3801 & 0.3937 & 0.5616 & \bfseries \color{red}0.3588 & \bfseries 0.3784 & \bfseries 0.3796 & \bfseries 0.3792  \\
\cline{2-9}
 & std & 0.0071 & 0.0046 & 0.0037 & 0.0039 & 0.0065 & 0.0074 & 0.0035 \\
\cline{1-9}

\multirow{2}{*}{NSGA-III} & Mean IGD & 0.4263 & 0.7102 & 0.5815 &  \bfseries \color{red}0.4210 & \bfseries 0.4238 & \bfseries 0.4234 & \bfseries 0.4250\\
\cline{2-9}
 & std & 0.0113 & 0.0213 & 0.0042 & 0.0093 & 0.0086 & 0.0086 & 0.0110 \\

\hline
\end{tabular}
\label{dtlz2-convex}
\end{table}

\begin{table}[!htbp]
\centering
\caption{The ranking of mean hypervolume.}
\begin{tabular}{|c|c|c|c|c|c|c|c|c|}
\hline

Problems & Algorithms & P\_cone & $\frac{\pi}{6}$(=30$^{\circ}$) & $\frac{\pi}{9}$(=20$^{\circ}$) & $\frac{\pi}{12}$(=15$^{\circ}$) & $\frac{\pi}{18}$(=10$^{\circ}$) &
$\frac{\pi}{30}$(=6$^{\circ}$) & $\frac{\pi}{60}$(=3$^{\circ}$) \\
\hline
DTLZ2 & NSGA-II & 5 & 7 & 6 & 1 & 2 & 3 & 4  \\
\cline{2-9}
(concave) &DI-MOEA & 5 & 7 & 6 & 1 & 2 & 3 & 4  \\
\cline{2-9}
 4 obj& NSGA-III & 4 & 7 & 6 & 1 & 2 & 3 & 5  \\
\cline{1-9}
\hline

DTLZ2 & NSGA-II & 6&	7&	5&	1&	2&	3&	4  \\
\cline{2-9}
(concave) &DI-MOEA & 6&	6&	5&	1&	2&	3&	4  \\
\cline{2-9}
6 obj & NSGA-III & 5	&7&	6&	1&	2&	3&	4  \\
\cline{1-9}
\hline

DTLZ2 & NSGA-II & 6&	7&	4&	1&	2&	3&	5 \\
\cline{2-9}
(concave) &DI-MOEA & 6&	6&	4&	1&	2	&3&	5 \\
\cline{2-9}
8 obj & NSGA-III & 4	&7&	6&	1&	2&	3&	5 \\
\cline{1-9}
\hline

\multirow{3}{*}{UF11} & NSGA-II & 6&	6&	4&	2&	1&	3&	5  \\
\cline{2-9}
 &DI-MOEA & 6&	7&	5&	1&	2&	3&	4  \\
\cline{2-9}
 & NSGA-III & 6	&7	&3&	1&	2&	4&	5  \\
\cline{1-9}
\hline

\multirow{3}{*}{UF13} & NSGA-II & 6&	7&	2&	1&	3&	4	&5 \\
\cline{2-9}
 &DI-MOEA & 3&	7&	1&	2&	4&	6&	5  \\
\cline{2-9}
 & NSGA-III &6&	7&	2&	1&	3&	4&	5  \\
\cline{1-9}
\hline

DTLZ1 & NSGA-II & 7&	6&	1&	2&	3&	5&	4  \\
\cline{2-9}
 &DI-MOEA & 6&	7&	3&	2&	5&	1&	4 \\
\cline{2-9}
4 obj & NSGA-III &5&	7&	4&	6&	1&	2&	3  \\
\cline{1-9}
\hline

DTLZ1 & NSGA-II & 7&	6&	1&	2&	3&	4&	5  \\
\cline{2-9}
 &DI-MOEA & 7&	6&	1&	1&	3&	4&	5  \\
\cline{2-9}
6 obj & NSGA-III & 6&	7&	1&	2&	3&	5&	3 \\
\cline{1-9}
\hline

DTLZ1 & NSGA-II & 7&	6&	1&	2&	3&	4&	5 \\
\cline{2-9}
 &DI-MOEA & 7&	6&	1&	1&	3&	4&	5 \\
\cline{2-9}
 8 obj& NSGA-III & 1&	7&	4&	3&	6&	2&	5 \\
\cline{1-9}
\hline

DTLZ2 & NSGA-II & 5&	7&	6&	1&	2&	3&	4 \\
\cline{2-9}
(convex) &DI-MOEA & 5&	7&	6&	4&	1&	2&	3 \\
\cline{2-9}
4 obj & NSGA-III &5&	7	&6	&2	&1&	3&	4  \\
\cline{1-9}
\hline

DTLZ2 & NSGA-II & 6&	7&	5&	1&	2&	3&	4 \\
\cline{2-9}
(convex) &DI-MOEA & 5&	7&	6&	1&	2&	3&	4 \\
\cline{2-9}
6 obj & NSGA-III &6&	7&	4&	1&	2&	3&	5  \\
\cline{1-9}
\hline

DTLZ2 & NSGA-II & 5&	7&	6&	1&	2&	3&	4 \\
\cline{2-9}
(convex) &DI-MOEA & 5&	6&	7&	1&	2&	3&	4 \\
\cline{2-9}
 8 obj& NSGA-III &4&	7&	6&	1&	2&	3&	5  \\
\cline{1-9}
\hline
\hline
\multicolumn{2}{|c|}{ ON AVERAGE}  & 5.4&	6.7	&4&	1.5&	2.4&	3.3	&4.4 \\
\cline{2-9}
\cline{1-9}
\hline

\multicolumn{9}{|c|}{\bfseries Half budget}\\
\hline
DTLZ2 & NSGA-II &6&	7&	4&	1&	2&	3&	5 \\
\cline{2-9}
(concave) &DI-MOEA & 6&	6&	4&	1&	2&	3&	5 \\
\cline{2-9}
8 obj & NSGA-III &5&	7&	6&	1&	2&	3&	4 \\
\cline{1-9}
\hline

\multirow{3}{*}{UF11} & NSGA-II &6&	6&	3&	2&	1&	4&	5\\
\cline{2-9}
 &DI-MOEA & 6&	7&	3&	1&	2&	4&	5 \\
\cline{2-9}
 & NSGA-III &6&	7&	3&	1&	2&	4&	5\\
\cline{1-9}
\hline

\multirow{3}{*}{UF13} & NSGA-II &6&	7&	1&	2&	3&	4&	5\\
\cline{2-9}
 &DI-MOEA & 6&	7&	1&	2&	3&	5&	4 \\
\cline{2-9}
 & NSGA-III &5&	7&	1&	2&	3&	4&	6\\
\cline{1-9}
\hline

\multirow{3}{*}{DTLZ1} & NSGA-II &6&	6&	1&	2&	3&	4&	5 \\
\cline{2-9}
 &DI-MOEA & 6&	6&	3&	1&	2&	4&	5 \\
\cline{2-9}
 & NSGA-III &6&	7&	3&	5&	4&	2&	1 \\
\cline{1-9}
\hline

DTLZ2 & NSGA-II &5&	7&	6&	1&	2&	3&	4 \\
\cline{2-9}
(convex) &DI-MOEA & 5&	6&	7&	1&	2&	3&	4\\
\cline{2-9}
8 obj & NSGA-III &4&	7&	6&	1&	2&	3&	5\\
\cline{1-9}
\hline
\hline
 \multicolumn{2}{|c|}{ON AVERAGE} & 5.6& 	6.7& 	3.46& 	1.6	& 2.3& 	3.53& 	4.5\\
\cline{2-9}
\cline{1-9}
\hline

\end{tabular}
\label{rank}
\end{table} 

\end{document}